\documentclass[runningheads]{llncs}

\usepackage{amsmath}
\usepackage{esvect}    % add this after amsmath
\usepackage[table]{xcolor}
\usepackage{multirow}
\usepackage{amsfonts}
\usepackage{graphicx}
\usepackage{anyfontsize}
\usepackage{soul}
\usepackage{iciap}
\usepackage[hyphens]{url}
\usepackage[
    colorlinks=true,
    pdfborder={0 0 0},
    linkcolor=red
]{hyperref}
\newcommand{\mycomment}[1]{}

\usepackage{booktabs} %\specialrule{2.5pt}{1pt}{1pt}
\usepackage[normalem]{ulem}
\useunder{\uline}{\ul}{}
\usepackage{pifont}
\usepackage[dvipsnames]{xcolor}
\newcommand{\xmark}{\ding{56}}

\def\debug{0}

\ifnum \debug=1
\providecommand{\lnote}[1]{\textcolor{blue}{[LR: #1]}}
\providecommand{\mnote}[1]{\textcolor{violet}{[MB: #1]}}
\providecommand{\anote}[1]{\textcolor{orange}{[MA: #1]}}
\providecommand{\pnote}[1]{\textcolor{red}{[AP: #1]}}
\else
\providecommand{\lnote}[1]{}
\providecommand{\mnote}[1]{}
\providecommand{\anote}[1]{}
\providecommand{\pnote}[1]{}
\fi

\begin{document}
\title{CoLoR-GAN: Continual Few-Shot Learning with Low-Rank Adaptation in Generative Adversarial Networks}
%\title{\textcolor{Red}{CoLoR-GAN}: \textcolor{Red}{C}ontinual Few-Sh\textcolor{Red}{o}t Learning with \textcolor{Red}{Lo}w-\textcolor{Red}{R}ank Adaptation in Generative Adversarial Networks}
\titlerunning{CoLoR-GAN}
\author{Munsif Ali \inst{1} 
\and
Leonardo Rossi \inst{1}
\and
Massimo Bertozzi \inst{1}}

%\author{Anonymous Authors}

%
\authorrunning{M. Ali et al.}
\institute{Department of Engineering and Architecture, University of Parma, Italy
\email{\{munsif.ali, leonardo.rossi, massimo.bertozzi\}@unipr.it}}
\maketitle              % typeset the header of the contribution

\begin{abstract}
Continual learning (CL) in the context of Generative Adversarial Networks (GANs) remains a challenging problem, particularly when it comes to learn from a few-shot (FS) samples without catastrophic forgetting.
Current most effective state-of-the-art (SOTA) methods, like LFS-GAN, introduce a non-negligible quantity of new weights at each training iteration, which would become significant when considering the long term.
For this reason, this paper introduces \textcolor{red}{\textbf{\underline{c}}}ontinual few-sh\textcolor{red}{\textbf{\underline{o}}}t learning with \textcolor{red}{\textbf{\underline{lo}}}w-\textcolor{red}{\textbf{\underline{r}}}ank adaptation in GANs named CoLoR-GAN, a framework designed to handle both FS and CL together, leveraging low-rank tensors to efficiently adapt the model to target tasks while reducing even more the number of parameters required.
Applying a vanilla LoRA implementation already permitted us to obtain pretty good results.
In order to optimize even further the size of the adapters, we challenged LoRA limits introducing a LoRA in LoRA (LLoRA) technique for convolutional layers.
Finally, aware of the criticality linked to the choice of the hyperparameters of LoRA, we provide an empirical study to easily find the best ones.
We demonstrate the effectiveness of CoLoR-GAN through experiments on several benchmark CL and FS tasks and show that our model is efficient, reaching SOTA performance but with a number of resources enormously reduced.
Source code is available on \href{https://github.com/munsifali11/CoLoR-GAN}{Github.}

\end{abstract}

\section{Introduction}

Recent advancements \cite{brock2018large,styleGAN,styleGAN2,song2022editing} in GANs show promising results on existing benchmark datasets. However, these models rely on a large amount of data for better performance.
Usually, when new data is available for training, the model updates all the parameters because a trains from scratch is needed to preserve old generation capabilities and extend it with the new data.
This is not always desirable, because it requires a long training and every time more longer, because the dataset increases during time.
Neither, it is a feasible behavior, because not always all the data are available.
In the worst case, as in the Continual Learning (CL) scenario, past data is not available anymore and training on new data only exposes the model to the phenomenon known as catastrophic forgetting~\cite{mccloskey1989catastrophic}.
%Moreover, the model overwrites the new parameters on older ones while training the model for new tasks, this phenomenon is known as catastrophic forgetting \cite{mccloskey1989catastrophic}. Therefore these models are very challenging to address these issues and introduce two aspects of FS and CL.
Moreover, in many scenarios due to privacy concerns and domains which have limited data such as health, art, etc., only a few samples are available for each task, making it challenging for models to adapt efficiently \cite{tian2024survey}.
Therefore, in Few-Shot (FS) the challenge arises from the few samples (e.g. $\leq 10$) available for training which can lead to over-fitting \cite{zheng2023my}. 
%On the other hand, CL is the ability of the model to learn new tasks without forgetting \cite{mccloskey1989catastrophic}. 

Unlike discriminative models, GANs are not explored very well in the context of both FS and CL. 
GAN is known for its ability to generate high-quality synthetic data but faces significant difficulties in handling the evolving nature of tasks. Therefore, considering both FS and CL together is an emerging and challenging area~\cite{lfs}. In this context, the GAN have to learn sequentially with few training samples over time without affecting the previous knowledge.

Existing studies such as regularization \cite{cdc,rssa,dcl} focus only on FS and fine-tune all the parameters during the adaptation. These approaches are very good to maintain the diversity of the source domain. However, they do not consider CL and that limit the applicability of generating previously learned representations. On the other hand, many methods \cite{ali2025cfts,gan_memory,lfs,zhai2021hyper} focus only on mitigating catastrophic forgetting by expanding the source model. Nevertheless, these models require a large number of parameters for the target tasks, leading to a storage burden to save the target parameters. A recent study~\cite{lfs} proposed a modulation-based efficient approach that considered both FS and CL together and produced high-quality and diverse images. Unfortunately, the adapted parameters are very complex and still need consideration to reduce them. Furthermore, a simple approach with less parameters can be effective for FS, as training with too many parameters can lead to over-fitting \cite{dong2025low}.

In response to these challenges, this paper introduces CoLoR-GAN (continual learning with low-rank adaptation in GANs). Our approach takes inspiration from LoRA \cite{lora} and appends low-rank parameters to the source model to address the issues of catastrophic forgetting in CL scenarios. Low-rank adaptation facilitates efficient parameter updates and reduces the risk of over-fitting to the target tasks, while enabling the model to effectively learn tasks without forgetting.
Furthermore, we extend LoRA methodology to reduce the parameters number in case of convolution, introducing a LoRA in LoRA mechanism.
Moreover, because LoRA is usually very sensitive to hyper-parameters selection, we describe an easy approach to select the most appropriate ones.
The evaluation of CoLoR-GAN on several benchmark FS datasets demonstrates its effectiveness in mitigating catastrophic forgetting while achieving comparable performance to SOTA methods but with less parameters and less training iterations.

The main contributions of our work are:  
\begin{itemize}
    \item Inspired by LoRA \cite{lora}, we added the low-rank tensor on top of StyleGAN2 for CL and FS image generation; introducing also a LoRA in LoRA (LLoRA) for convolutions, in order to even further decrease the number of parameters.\mnote{maybe stress this} \anote{also have look at it}
    \item We provide an empirical study to choose the  most effective LoRA hyper-parameters in generative CL and FS tasks.
    \item We evaluated the CoLoR-GAN through experiments on the CL and FS datasets and demonstrated its performance in terms of quality and diversity compared the state-of-the-art.

\end{itemize}

\section{Literature Review}
\noindent \textbf{FS Learning:}
In FS, we are in a situation where data is scarce and we have only a few samples to train a model \cite{tgan}. 
Recent advancements leverage ensemble models~\cite{gal2022stylegan,sauer2021projected}, incorporating techniques such as regularization~\cite{ewc,hou2023regularizing}, expanding network \cite{adam}, or contrastive learning \cite{dcl,lee2021c3} to enhance performance in FS.  
\noindent The CDC \cite{cdc} introduces a method for FS image generation by leveraging cross-domain correspondence loss. RSSA \cite{rssa} introduces a relaxed spatial structural consistency loss enables the model to maintain structural information from source images. In \cite {dcl} the DCL method is proposed with the aim of retaining the diversity of the source domain using the contrastive loss. AdAM \cite{adam} is a modulation-based approach. Instead of fine-tuning all the parameters, it introduces additional modulation parameters that handle FS learning, while keeping the source parameters frozen. 
All these approaches work very well. However, they focus solely on FS learning. In contrast, our method addresses the more challenging task of combining FS learning with CL.

\noindent \textbf{CL:} CL in GANs research area addresses the challenge of maintaining knowledge across multiple tasks. Various methods, including rehearsal strategies \cite{wu2018memory}, regularization techniques \cite{ewc}, and architectural expansion \cite{cam_gan,verma2021efficient} exist for CL in GANs. 
The CAM-GAN \cite{cam_gan} introduces a novel CL framework for  GANs. The approach leverages adapter modules integrated into the existing architecture \cite{gp_gan}. The adapters ensure the learning of new tasks and adapt incrementally. An alternative approach for both discriminative and generative models is presented in \cite{verma2021efficient}, introducing efficient parameters for CL. Recently, low-rank adaptation (LoRA) \cite{lora} gained popularity in LLMs for CL \cite{zheng2024towards}. It reduces the memory and computation overhead when adapting a pre-trained model to new tasks. A recent approach utilizing LoRA for diffusion models is proposed in \cite{wistuba2023continual}. Although these methods achieve promising results, they do not address FS learning. In contrast, our model considers a more challenging task of FS and CL, offering a distinct advancement over previous approaches.

\noindent \textbf{CL and FS Learning:}
One of the most challenging task considers both FS and CL together. In this setup, models adapt to new tasks with only a few training samples while simultaneously maintaining the performance on previously learned tasks \cite{mccloskey1989catastrophic}. 
The approach for conditional GANs in \cite{le2023mode} leverages the replay method and a novel discriminative mode affinity loss to address CL and FS. 
Another approach introduced in \cite{ali2025cfts} for both FS and CL considers the teacher-student model with CDC loss. 
A recent approach named LFS-GAN \cite{lfs} presents an approach that considered both FS and CL in GANs.
This method introduces low-rank tensors to learn new tasks while retaining previously acquired knowledge efficiently. Additionally, the approach incorporates a loss function to enhance the diversity of synthesized images.
These approaches provide the best results. However, they require more parameters and training loops compared to ours. 
Unlike LFS-GAN, which relied on complex operations (i.e. decomposition and reconstruction of multiple tensors and modulates the pre-trained parameters), we took inspiration from~\cite{lora} and incorporated simple low-rank tensors and added them to pre-trained weights. In this way, our model gives comparable results and maintains the source information while enabling more efficient training with significantly fewer parameters.

\section{CoLoR-GAN model}

\subsection{Preliminaries}
\textbf{LoRA Background:}
Because fully training a Large Language Model (LLM) is very expensive, the LoRA \cite{lora} finetuning trick has been proposed to reduce, as much as possible, the number of weights involved in the training without performance penalties.
The low-rank adaptation incrementally updates internal pre-trained model weights.
Consider a pre-trained weight matrix $W_0 \in \mathbb {R}^{d \times k}$. Instead of modifying the entire weight matrix during incremental fine-tuning, it approximates the update as a low-rank adaptation of two matrices $B \in \mathbb{R}^{d \times r}$ and $A \in \mathbb{R}^{r \times k}$, where $r$ is the rank of the matrix. The adapted weights $\hat{W}$ are given as follows:
\begin{equation}\label{lora_eq}
\hat{W} = W_0 +  \Delta W=W_0 + \frac{\alpha}{r}BA 
\end{equation}
\noindent where $ \Delta W$ represents the weight updates and $\alpha$ the scaling factor that controls the LoRA influences.  
%Rather than storing the full matrix, LoRA decomposes it into the product of two low-rank matrices. 
%The underlined term represents the weights that need to be fine-tuned for the downstream tasks.
Selecting an $r$ much lower than $d$ and $k$, means that $d\times r + r\times k \ll d\times k$ and therefore that the additional trained parameters can be really small.

\noindent\textbf{CL and FS Definition:}
In the case of continual learning, different tasks $T = {T_1, T_2, ... T_n}$ are considered to train the model while maintaining performance on previously learned tasks without revisiting them. Let $D_i = \left\{ x_i^t \right\}_{i =1}^{k}$ be the dataset for the task $T_i$. 
In FS case, the $k$ is limited to a very small number and represents the number of training samples in the dataset for the task $T_i$. 
At time $i$, only the current task is available to train the model. Let $W_i$ represents the model parameters after learning task $T_i$. The goal of continual learning is to optimize the weights on each task sequentially and preserves the ability to generates all the tasks.

\subsection{StyleGAN2 LoRA Adaptation}

\begin{figure}[t!]
    \centering
    \includegraphics[width=0.62\textwidth]{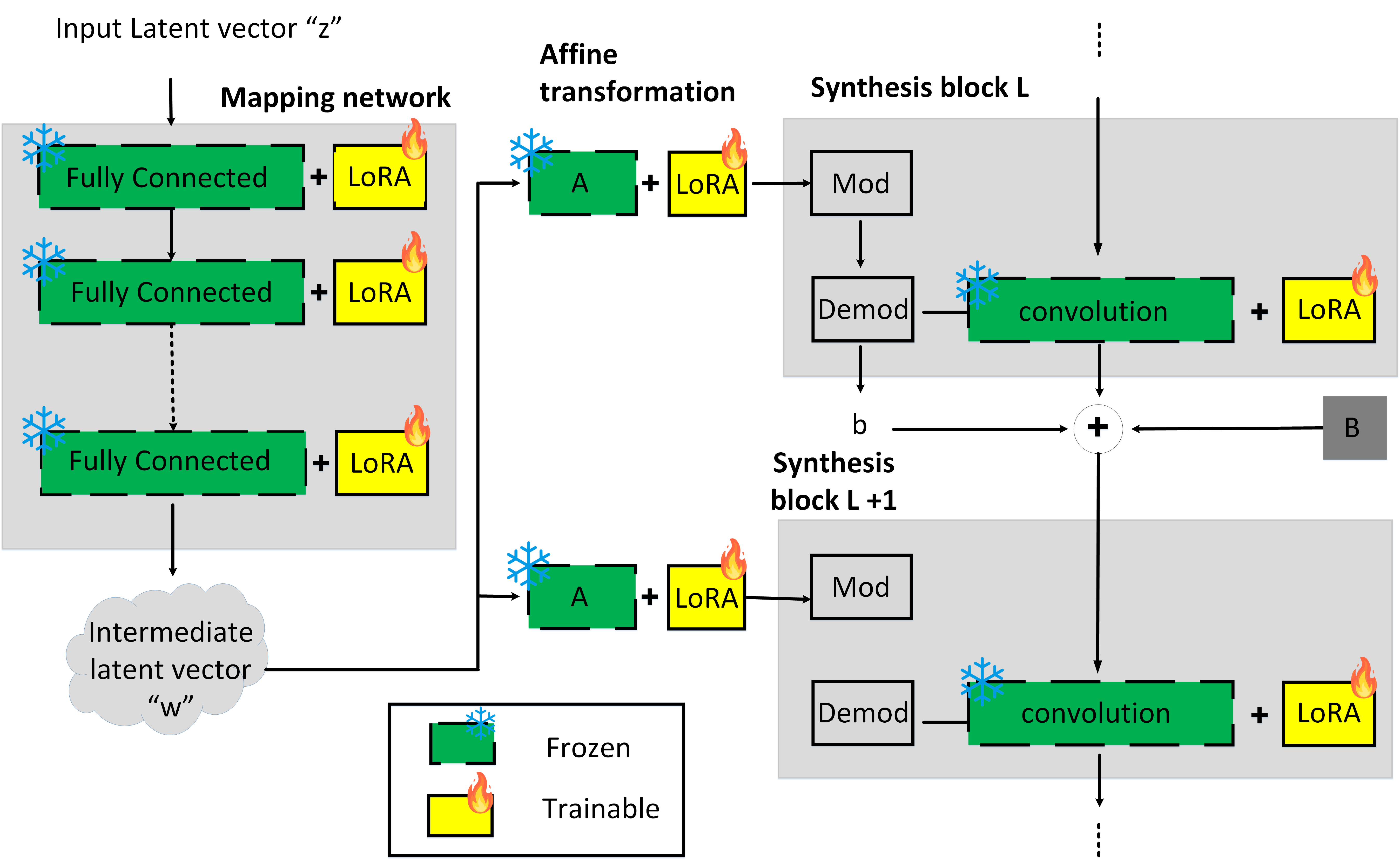}
    \caption{Low-rank adaptation (solid yellow) is incorporated into the pretrained StyleGAN2 model (solid green) for continual few-shot image generation. For a CL FS task, only the adapted weights are trained. Block A and B are the affine transformation and the injected noise used in the original StyleGAN2, respectively. $w$ represents the intermediate latent vector.}
    \label{lora_model}
\end{figure}

Let's revisit the original StyleGAN2 model, which is very popular for synthesizing images. It consists of two modules: mapping and synthesis network. The first one translates a noise into a latent vector, while the synthesis network transforms the intermediate latent vector into an image. In Fig.~\ref{lora_model}, the StyleGAN2 is shown with dashed green boxes.
\noindent On the top of StyleGAN2, we adapted the low-rank tensor as given in Eq.~\ref{lora_eq} into the mapping, and synthesis modules.
The LoRA adaptation for CL and FS generation is shown in Fig.~\ref{lora_model} with solid yellow boxes.
The pretrained StyleGAN2 weights are frozen while only the low-rank tensors are trained for the downstream tasks. 
By freezing the pretrained weights of StyleGAN2 and training only the low-rank tensors, we aim to adapt the model to new tasks while leveraging and retaining the source model's knowledge. 
Each new task involves new adapted trained weights, which are preserved to save past generation abilities when new tasks are available.

\noindent \textbf{LoRA FC:} \label{lora_fc_text}
In the mapping network, we extended each fully connected (FC) layer with a LoRA FC layer.
An FC layer is composed of a weight tensor $W_{FC} \in  \mathbb{R}^{d_{out} \times d_{in}}$.
LoRA FC introduces $W_{FC}$, composed by two new low rank tensors $B \in \mathbb{R}^{d_{out} \times r}$ and $A \in \mathbb{R}^{r \times d_{in}}$: 
\begin{equation}
\begin{split}
\Delta W &= B \times A \\
\hat{W}_{FC} &= W_{FC} + \frac{\alpha_{fc}}{r}{\Delta W}
\end{split}
\end{equation}
\noindent where $d_{in}$ and  $d_{out}$ are the input and output dimensions, $r$ represents the rank, $\hat{W}_{FC}$ represents the new FC weights after LoRA adaptation and the symbol $\times$ represents matrix multiplication.

\noindent \textbf{LLoRA Conv:} \label{lora_conv_text} 
In the synthesis part, we extended all convolutional layers with a corresponding LoRA module.
Due to the high dimension of the convolution weight tensor $W_{conv} \in  \mathbb{R}^{c_{out} \times c_{in} \times k \times k}$, we define a LoRA in LoRA (LLoRA) adapter.
As previously done for FC, we define the new convolutional weights as:
\begin{equation}
\begin{split}
  \Delta W_{conv} &= act(B \times A) \\
  \hat{W}_{conv} &= W_{conv} + \frac{\alpha_{conv}}{r}{\Delta W_{conv}}
\end{split}
\end{equation}
\noindent where $W_{conv}$, $\hat{W}_{conv}$ and $\Delta W_{conv}$ represent the pretrained frozen convolutional weights, the convolutional weights after LoRA adaptation and the LoRA adaptation weights, respectively.
All of them have the shape $\mathbb{R}^{c_{out} \times c_{in} \times k \times k }$.
And, $\alpha$, $c_{in}$, $c_{out}$, and $k \times k$ are the hyper-parameter to control the target adaptation, the in and out channels, and the kernel size, respectively.

The $B \in \mathbb{R}^{r \times c_{out} \times k \times k}$ and $A \in \mathbb{R}^{r \times c_{in}}$ weights represents the low-rank matrices that are learned for each task.
Here, differently from LoRA FC, to enhance model expressivity, we also introduce a non-linearity to compute $\Delta W_{conv}$ (a ReLU activation function), because it plays a crucial role in significantly enhancing the model's performance and expressivity. 

\noindent Because the number of weights in $B$ could be relatively high, to reduce its impact we take inspiration from \cite{lfs}, and factorize it in the following way:
\begin{equation}
\begin{split}
B &= act(B' \times M_{inst}) \\
\end{split}
\end{equation}
\noindent where $B' \in \mathbb{R}^{c_{out} \times r}$ and $M_{inst} \in \mathbb{R}^{r \times r \times k \times k}$ are the low rank matrices that multiplied together are forming the $B$.
Because this low-rank parametrization is a LoRA itself, and is positioned inside a LoRA, we define this architecture as LoRA in LoRA (LLoRA) Conv.
It is worth nothing that another non-linearity is introduced to enhance model expressivity even further.

\noindent \textbf{Scaling factor}:\label{alpha_text}
The constant value $\frac{\alpha}{r}$ represents a scaling factor that influences how much the LoRA parameters weigh over the total.
Usually, the divisor is set equal to the rank $r$, and there is no general rule to choose the dividend $\alpha$.
In literature, the choice of this value was explored in other contexts \cite{biderman2024lora,rslora,tribes2023hyperparameter}, leading to empirical choices.
As noted by the authors, final performance is very sensitive to this value.
We tested that, automatically learning the value does not always pay.
We measured the distance between target and source ($L_{s-t}$), computing the average of pairwise LPIPS (Learned Perceptual Image Patch Similarity) \cite{lpips} distance within each cluster. 
Empirically, we noted that for target domains far from the dataset on which the model was originally pretrained (source dataset), bigger $\alpha_{fc}$ for LoRA FC and smaller $\alpha_{conv}$ for LLoRA Conv work better.
We suppose that it is because the FC layer handles a more abstract representation, therefore, it benefits from larger adaptation while the convolution layers capture more spatial features.
From the experiments, we notice that the higher the LPIPS distance from the source, higher $\alpha_{fc}$ than 1 and the lower $\alpha_{conv}$ than 1 work better.
In most cases, best $\alpha_{fc}$ is around 1.5 and best $\alpha_{conv}$ around 0.25.
More the $L_{s-t}$ is higher and more this is true.
In other words, for a dataset far from the source domain, the LoRA weight for the Mapping network need to be amplified and for the Synthesis network need  to be attenuated.

\section{Experiments}
\noindent\textbf{Implementation:}
We followed state-of-the-art \cite{lfs} and utilized the pre-trained StyleGAN2 \cite{styleGAN2} on the FFHQ \cite{styleGAN} dataset and adopted the low-rank tensors for the target domain generation on the top of StyleGAN2 \cite{styleGAN2}. The pre-trained model remains frozen, while only low-rank tensors are trained to adapt to the target domains. The model is trained using Adam optimizer with a learning rate of 0.002 and a batch size of 4 due to few-shot learning, ensuring efficient updates and steady convergence \cite{lfs}, the training is stopped at 1500 iterations.
The loss used to train the GAN is a Wasserstein Loss \cite{arjovsky2017wasserstein}.

\noindent \textbf{Datasets:}
The experiments are carried out on multiple datasets. The StyleGAN2 pre-trained on FFHQ \cite{styleGAN} and the  LSUN-Car \cite{yu2015lsun} are considered as a source domain. The target datasets include Sketches, Females, Sunglasses, Males, Babies \cite{deng2009imagenet,cdc}, wrecked car\footnote{The link for wrecked car dataset is \href{https://www.kaggle.com/datasets/prajwalbhamere/car-damage-severity-dataset/data}{here}}, and truck datasets \footnote{The link for truck dataset is \href{https://zenodo.org/records/5744737}{here}}. Each target dataset includes 10 samples for training. These diverse target datasets help assess the performance of our model across different target domains.

\noindent \textbf{Baselines:}
We considered several state-of-the-art models for comparison that address three distinct aspects of image generation: (i) FS image generation, (ii) CL image generation, and (iii) FS and CL together. The methods CDC \cite{cdc}, RSSA \cite{rssa}, DCL \cite{dcl}, and AdAM \cite{adam} belong to the first category, focusing on FS image generation. In contrast, GAN-Memory \cite{gan_memory} and CAM-GAN \cite{cam_gan} are designed to handle CL. While the LFS-GAN \cite{lfs} considered both FS and CL, similar to ours. 

\noindent \textbf{Metrics:}
The FID \cite{heusel2017gans} and B-LPIPS \cite{lfs} metrics are used to assess the performance of our model. The FID score measures the similarity between synthesized and real images, indicating that a lower score corresponds to better performance. 
On the other hand, the B-LPIPS metric evaluates the diversity of the synthesized images, and a higher value represents better results.  To evaluate FID, we sampled $5000$ images, while $1000$ samples are used for B-LPIPS. We followed the same training and evaluation protocol and datasets of LFS-GAN to be as fair as possible.

\subsection{Few-shot and Continual Learning Evaluation}

\begin{figure}[t!]
    \centering
    \includegraphics[width= 0.49\textwidth]{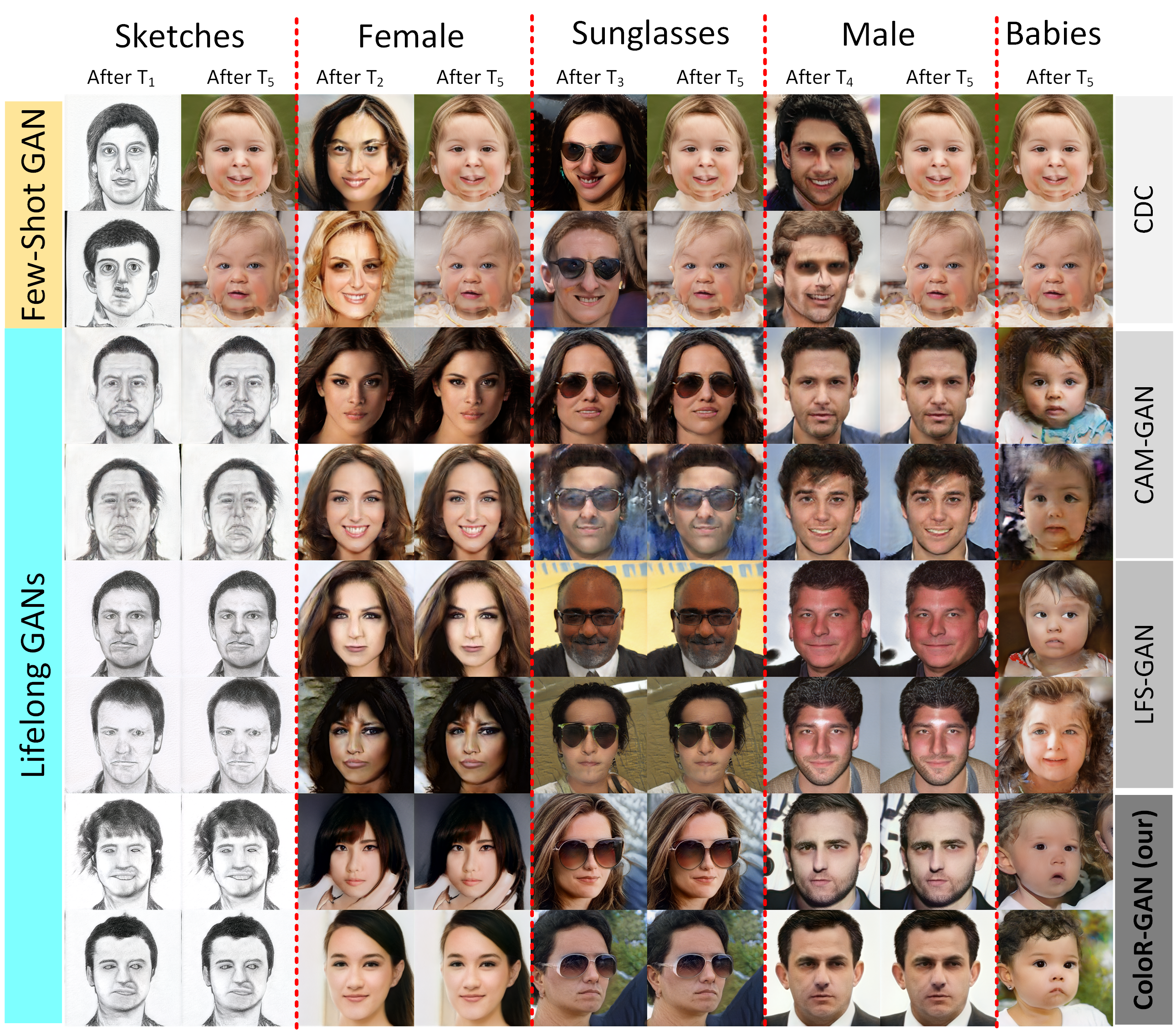}
    %\caption{Qualitative results comparison. The few-shot GAN is not able to produce the previous samples, while lifelong GANs are able to avoid catastrophic forgetting. Each pair of rows shows two samples from each domain.}
    \caption{Qualitative results. Comparison of continual image generation of different models. The few-shot GAN is not able to produce the previous samples while lifelong GANs are able to avoid catastrophic forgetting. Each pair of rows shows two samples from each domain.}
    \label{img:quantitive_results_cl}
\end{figure}

\noindent \textbf{Qualitative Results:}
First of all, the CoLoR-GAN model is evaluated for the FS and CL task. This means that, after training the model on the current FS dataset, all previous tasks are evaluated as well to check that the model has not forgotten the past. 
As shown in Fig.~\ref{img:quantitive_results_cl}, our model effectively generates all the previous tasks and maintains quality and performance without affecting the previous information. 
It demonstrates that CoLoR-GAN is able to keep the learned information from the earlier tasks without degradation or interference like the compared state-of-the-art model.

\noindent \textbf{Quantitative Results:} 
As shown in Table \ref{tab:quantitive_results}, TGAN and other FS methods fail to generate quality samples due to their lack of consideration for CL. Therefore, the FID is very higher for these methods. Among FS methods, AdAM performs better than the others because it does not update all the weights during adaptation.
On the other hand, CL methods such as GAN-Memory and CAM-GAN yield better results than FS methods by incorporating CL. However, these methods do not account for the FS scenario, leading them to replicate the training data and over-fit. \anote{does not ref anymore to supplementary. is it OK?} 
%as shown in Fig.~\ref{img:sunglasses_fs}.
In contrast, our model and LFS-GAN effectively address both FS and CL, producing high-quality and diverse images. 
Despite LFS-GAN strengths, there is still room for improvement, because our model CoLoR-GAN achieves almost the same performance with half parameters and half iterations.
The training efficiency, measured by the number of trainable parameters and iterations, is shown in Table \ref{tab:training_efficiency}.

\begin{table}[t!]
\centering
\fontsize{4.3pt}{5.5pt}\selectfont
\caption{\textbf{Continual few-shot image generation}: comparison with different methods in terms of FID ($\downarrow$) and B-LPIPS ($\uparrow$) across different tasks. The results of all models are taken after training the model on the last (Babies) task. \textcolor{red}{\textbf{Best}} and \underline{\color[HTML]{6200C9}{2nd highest}} results.}

\begin{tabular}{ll|cc|cc|cc|cc|cc|ll}
\toprule \hline
\cellcolor[HTML]{FFFFFF}Methods & \cellcolor[HTML]{FFFFFF} & \multicolumn{2}{c|}{\cellcolor[HTML]{60E7F5}\textbf{Sketches ($T_1$)}} & \multicolumn{2}{c|}{\cellcolor[HTML]{60E7F5}\textbf{Female ($T_2$)}} & \multicolumn{2}{c|}{\cellcolor[HTML]{60E7F5}\textbf{Sunglasses ($T_3$)}} & \multicolumn{2}{c|}{\cellcolor[HTML]{60E7F5}\textbf{Male ($T_4$)}} & \multicolumn{2}{c|}{\cellcolor[HTML]{60E7F5}\textbf{Babies ($T_5$)}} & \multicolumn{2}{c}{\cellcolor[HTML]{60E7F5}\textbf{Average}} \\ \cline{3-14} 
 &  & \cellcolor[HTML]{C0C0C0}FID & \cellcolor[HTML]{C0C0C0}B-LPIPS & \cellcolor[HTML]{C0C0C0}FID & \cellcolor[HTML]{C0C0C0}B-LPIPS & \cellcolor[HTML]{C0C0C0}FID & \cellcolor[HTML]{C0C0C0}B-LPIPS & \cellcolor[HTML]{C0C0C0}FID & \cellcolor[HTML]{C0C0C0}B-LPIPS & \cellcolor[HTML]{C0C0C0}FID & \cellcolor[HTML]{C0C0C0}B-LPIPS & \cellcolor[HTML]{C0C0C0}FID & \cellcolor[HTML]{C0C0C0}B-LPIPS \\ \hline
\multicolumn{2}{c|}{Baseline (TGAN) \cite{tgan}} & 372.89 & 0.157 & 255.33 & 0.238 & 309.13 & 0.247 & 281.43 & 0.129 & 171.19 & 0.203 & \multicolumn{1}{c}{278.02} & 0.195 \\ \hline
\multicolumn{1}{l|}{Few-shot GANs} & \multicolumn{1}{l|}{CDC \cite{cdc}} & 322.72 & 0.205 & 197.40 & 0.427 & 244.94 & 0.463 & 277.00 & 0.381 & 69.98 & 0.454 & 208.41 & 0.386 \\
\multicolumn{1}{l|}{} & \multicolumn{1}{l|}{RSSA \cite{rssa}} & 308.00 & 0.285 & 175.20 & 0.440 & 207.58 & 0.484 & 205.49 & 0.405 & 76.70 & 0.481 & 194.59 & 0.419 \\
\multicolumn{1}{l|}{} & \multicolumn{1}{l|}{DCL \cite{dcl}} & 297.73 & {\ul \color[HTML]{6200C9} 0.307} & 170.31 & 0.435 & 191.54 & 0.490 & 194.42 & {\ul \color[HTML]{6200C9} 0.443} & 77.22 & 0.487 & 186.25 & 0.432 \\
\multicolumn{1}{l|}{} & \multicolumn{1}{l|}{Adam \cite{adam}} & 161.48 & 0.250 & 179.69 & 0.342 & 217.19 & 0.352 & 163.87 & 0.299 & 110.08 & 0.407 & \multicolumn{1}{c}{166.82} & 0.330 \\ \hline
\multicolumn{1}{l|}{Continual} & \multicolumn{1}{l|}{GAN-Memory \cite{gan_memory}} & 69.58 & 0.311 & 71.56 & 0.287 & 87.02 & 0.169 & 99.44 & 0.143 & 177.73 & 0.150 & \multicolumn{1}{c}{101.05} & 0.212 \\
\multicolumn{1}{l|}{learning GANs} & \multicolumn{1}{l|}{CAM-GAN \cite{cam_gan}} & 91.81 & 0.293 & 85.68 & 0.332 & 86.81 & 0.333 & 82.83 & 0.312 & 146.20 & 0.181 & 98.66 & 0.290 \\ \hline
\multicolumn{1}{l|}{Continual and} & \multicolumn{1}{l|}{LFS-GAN \cite{lfs}} & \textcolor{red}{\textbf{34.66}} & \textcolor{red}{\textbf{0.354}} & \textcolor{red}{\textbf{29.59}} & \textcolor{red}{\textbf{0.481}} & \color[HTML]{6200C9} {\ul 27.69} & \textcolor{red}{\textbf{0.584}} & {\ul \color[HTML]{6200C9} 35.44} & \textcolor{red}{\textbf{0.472}} & \textcolor{red}{\textbf{41.48}} & \textcolor{red}{\textbf{0.556}} & \textcolor{red}{\textbf{33.77}} & \textcolor{red}{\textbf{0.489}} \\
\multicolumn{1}{l|}{Few-shot GANs} & \multicolumn{1}{l|}{ours} & {\ul \color[HTML]{6200C9} 45.03} & 0.281 & {\ul \color[HTML]{6200C9} 35.30} & \color[HTML]{6200C9}{\ul{0.448}} & \textcolor{red}{\textbf{25.93}} & {\ul \color[HTML]{6200C9} 0.493} & \textcolor{red}{\textbf{34.40}} & 0.422 & {\ul \color[HTML]{6200C9} 43.83} & {\ul \color[HTML]{6200C9} 0.526} & {\ul \color[HTML]{6200C9} 36.89} & {\ul \color[HTML]{6200C9} 0.434} \\ \bottomrule
\end{tabular}
\label{tab:quantitive_results}
\end{table}

\begin{table}[t!]
\centering
\fontsize{4.5pt}{5.5pt}\selectfont
\caption{Comparison on training efficiency. TPs, ITs: number of trainable parameters and number of iterations to learn a new task for each method, respectively. The second column represents the percentage with respect to StyleGAN2.}
\begin{tabular}{ll|l|l|l}
\rowcolor[HTML]{60E7F5} \toprule \hline \textbf{\tiny Method} &  & \textbf{\begin{tabular}[c]{@{}l@{}}\tiny TPs (M)\end{tabular}} & \textbf{\begin{tabular}[c]{@{}l@{}}\tiny \% w.r.t.~\\backbone\end{tabular}} & \textbf{\begin{tabular}[c]{@{}l@{}}\tiny ITs\end{tabular}} \\ \hline
\multicolumn{2}{c|}{Baseline (TGAN) \cite{tgan}} & 30.0 & 100\% & - \\ \hline
\multicolumn{1}{l|}{\multirow{4}{*}{Few-shot GANs}} & CDC \cite{cdc} & 30.0 & 100\% & 5K \\
\multicolumn{1}{l|}{} & RSSA \cite{rssa} & 30.0 & 100\% & - \\
\multicolumn{1}{l|}{} & DCL \cite{dcl} & 30.0 & 100\% & 3K \\
\multicolumn{1}{l|}{} & AdAM \cite{adam} & 18.9 & 63.0\% & - \\ \hline
\multicolumn{1}{l|}{Continual Learning GANs} & GAN-Memory \cite{gan_memory} & 5.3 & 17.7\% & - \\
\multicolumn{1}{l|}{} & CAM-GAN \cite{cam_gan} & 2.3 & 7.70\% & - \\ \hline
\multicolumn{1}{l|}{Continual and Few-shot GANs} & LFS-GAN \cite{lfs} & 0.1 & 0.36\% & 3K \\
\multicolumn{1}{l|}{} & our & \textcolor{red}{\textbf{0.05}} & \textcolor{red}{\textbf{0.18\%}} & \textcolor{red}{\textbf{1.5K}} \\ \bottomrule
\end{tabular}
\label{tab:training_efficiency}
\end{table}

\subsection{Ablation Study:}

\begin{table}[t!]
\centering
\fontsize{4.3pt}{5.5pt} \selectfont
\caption{Ablation study for the rank for our LoRA and LLoRA.}
\begin{tabular}{c|ccc|cc|cc|cc|cc|ll}
\toprule
\hline
\multicolumn{1}{l}{} & \multicolumn{1}{l}{} & \multicolumn{2}{c|}{\cellcolor[HTML]{60E7F5}\textbf{Sketches ($T_1$)}} & \multicolumn{2}{c|}{\cellcolor[HTML]{60E7F5}\textbf{Female ($T_2$)}} & \multicolumn{2}{c|}{\cellcolor[HTML]{60E7F5}\textbf{Sunglasses ($T_3$)}} & \multicolumn{2}{c|}{\cellcolor[HTML]{60E7F5}\textbf{Male ($T_4$)}} & \multicolumn{2}{c|}{\cellcolor[HTML]{60E7F5}\textbf{Babies ($T_5$)}} & \multicolumn{2}{c}{\cellcolor[HTML]{60E7F5}\textbf{Average}} \\ \cline{3-14} 
rank & \# of Trainable Param. & \cellcolor[HTML]{C0C0C0}FID & \cellcolor[HTML]{C0C0C0}B-LPIPS & \cellcolor[HTML]{C0C0C0}FID & \cellcolor[HTML]{C0C0C0}B-LPIPS & \cellcolor[HTML]{C0C0C0}FID & \cellcolor[HTML]{C0C0C0}B-LPIPS & \cellcolor[HTML]{C0C0C0}FID & \cellcolor[HTML]{C0C0C0}B-LPIPS & \cellcolor[HTML]{C0C0C0}FID & \cellcolor[HTML]{C0C0C0}B-LPIPS & \cellcolor[HTML]{C0C0C0}FID & \cellcolor[HTML]{C0C0C0}B-LPIPS \\ \hline
1 & \multicolumn{1}{c|}{{\color[HTML]{FF0000} {\textbf{54K}}}} & {\color[HTML]{FF0000} {\textbf{45.56}}} & {\color[HTML]{FF0000} {\textbf{0.280}}} & 35.30 & 0.448 & {\color[HTML]{FF0000} { \textbf{25.93}}} & {\color[HTML]{FF0000} { \textbf{0.493}}} & {\color[HTML]{FF0000} { \textbf{36.19}}} & 0.422 & {\color[HTML]{FF0000} {\textbf{47.20}}} & {\color[HTML]{FF0000} { \textbf{0.527}}} & {\color[HTML]{FF0000} { \textbf{37.91}}} & {\color[HTML]{FF0000} { \textbf{0.431}}} \\
2 & \multicolumn{1}{c|}{108K} & 50.59 & 0.163 & {\color[HTML]{FF0000} { \textbf{29.96}}} & {\color[HTML]{FF0000} { \textbf{0.449}}} & 32.66 & 0.482 & 37.77 & {\color[HTML]{FF0000} {\ \textbf{0.439}}} & 52.38 & 0.430 & 40.62 & 0.306 \\
4 & \multicolumn{1}{c|}{216K} & 50.64 & 0.188 & 30.52 & 0.415 & 43.29 & 0.475 & 37.12 & 0.399 & 53.82 & 0.526 & 43.07 & 0.295 \\
8 & \multicolumn{1}{c|}{432K} & 49.74 & 0.110 & 36.75 & 0.355 & 58.40 & 0.468 & 36.32 & 0.424 & 59.26 & 0.470 & 48.09 & 0.365 \\
\bottomrule
\end{tabular}
\label{tab:ablation_rank}
\end{table}

\noindent \textbf{Rank of adapted tensors:} 
In Table \ref{tab:ablation_rank}, we analyze the model on different ranks $r$ for all the datasets.
For most of the case, rank 1 provides in average the best results.
This result was predictable if we consider that we are in a limited-data regime.
The limited number of weights to be trained allows to take under control the overfitting effects.
It worth to note that, on female dataset, the rank 2 increases the number of parameters to the level of LFS-GAN, but delivering also the same performance.

\noindent \textbf{Activation function:}
We evaluated the model's performance with respect to the activation function.
As demonstrated in Table \ref{tab:ablation_act}, they plays a significant role in enhancing the performance. 

\noindent \textbf{Scaling factor:} 
Table \ref{tab:alpha_dist} shows results for the ablation on $\alpha$ value.
%Model with learnable $\alpha$ (first row) does not reach the best performance, although it always approaches the best value. \lnote{Is it better to remove alpha learnable?}
Similar to~\cite{eva}, we used fixed values for alpha in both parts.
On the bottom of the Table~\ref{tab:alpha_dist}, we show LPIPS $L_{s-t}$ distance between the Task $T_i$ and the original dataset FFHQ, on which the StyleGan2 was pretrained.
This LPIPS value is then used to select $\alpha_{fc}$ and $\alpha_{conv}$.
It worth noting that, more the Task $T_i$ dataset differs from FFHQ the bigger $\alpha_{fc}$ works better for FC, and the smaller $\alpha_{conv}$ works better for convolutions.
Best performances are usually obtained by assigning $\alpha_{fc} > 1$ ($\approx 1.5$) and $\alpha_{conv} < 1 $ ($\approx 0.25$). %\hl{ The beta threshold is selected empirically, and we set it to 0.4.}

%\noindent \textbf{LoRA https://www.ce.unipr.it/EEIV//extra-ue.phpFC and Conv:}
\noindent \textbf{LoRA FC and Conv:}
Furthermore, the adapted tensors in both parts are analyzed separately.
We compare generation performance of our model with only LoRA FC or with only LLoRA Conv.
As shown in Table \ref{tab:ablation_fc_conv}, it is clear that it performs more effectively when we add LoRA in both parts.

\noindent\textbf{Additional Experiments:} Taking the LSUN-Car as a source, we also performed experiments on the wrecked car and truck datasets.  Our model generates comparable results to LFS-GAN. 
It worth to noting that, our model beats LFS-GAN in the truck images, as shown in the right part of Table \ref{tab:car} , using $\alpha_{fc}$ of 1.626 and $\alpha_{conv}$ of 0.1, confirming our intuition.

\begin{table}[t!]
\centering
\fontsize{4.9pt}{5.5pt} \selectfont
\caption{Ablation study for the activation function.}
\begin{tabular}{c|cc|cc|cc|cc|cc|ll}
\toprule \hline
\multicolumn{1}{l|}{} & \multicolumn{2}{c|}{\cellcolor[HTML]{60E7F5}\textbf{Sketches ($T_1$)}} & \multicolumn{2}{c|}{\cellcolor[HTML]{60E7F5}\textbf{Female ($T_2$)}} & \multicolumn{2}{c|}{\cellcolor[HTML]{60E7F5}\textbf{Sunglasses ($T_3$)}} & \multicolumn{2}{c|}{\cellcolor[HTML]{60E7F5}\textbf{Male ($T_4$)}} & \multicolumn{2}{c|}{\cellcolor[HTML]{60E7F5}\textbf{Babies ($T_5$)}} & \multicolumn{2}{c}{\cellcolor[HTML]{60E7F5}\textbf{Average}} \\ \cline{2-13} 
Activation & \cellcolor[HTML]{C0C0C0}FID & \cellcolor[HTML]{C0C0C0}B-LPIPS & \cellcolor[HTML]{C0C0C0}FID & \cellcolor[HTML]{C0C0C0}B-LPIPS & \cellcolor[HTML]{C0C0C0}FID & \cellcolor[HTML]{C0C0C0}B-LPIPS & \cellcolor[HTML]{C0C0C0}FID & \cellcolor[HTML]{C0C0C0}B-LPIPS & \cellcolor[HTML]{C0C0C0}FID & \cellcolor[HTML]{C0C0C0}B-LPIPS & \cellcolor[HTML]{C0C0C0}FID & \cellcolor[HTML]{C0C0C0}B-LPIPS \\ \hline
\xmark & 57.32 & 0.238 & 40.74 & \color[HTML]{333333} 0.424 & 50.00 & \color[HTML]{000000} 0.457 & \color[HTML]{333333} 45.52 & \color[HTML]{333333} 0.421 & 70.76 & \color[HTML]{000000} 0.463 & \multicolumn{1}{c}{52.86} & \color[HTML]{333333} 0.400 \\
\checkmark & \color[HTML]{FF0000} \textbf{45.56} & \textbf{\color[HTML]{FF0000} 0.280} & \textbf{\color[HTML]{FF0000} 35.30} & \textbf{\color[HTML]{FF0000} 0.448} & \textbf{\color[HTML]{FF0000}25.93} & \cellcolor[HTML]{FFFFFF}{\color[HTML]{FF0000} \textbf{0.493}} & \textbf{\color[HTML]{FF0000} 36.19} & \textbf{\color[HTML]{FF0000} 0.422} & \textbf{\color[HTML]{FF0000}47.20} & \textbf{\color[HTML]{FF0000} 0.527} & \textbf{\color[HTML]{FF0000}37.91} & \textbf{\color[HTML]{FF0000} 0.431} \\ \bottomrule
\end{tabular}
\label{tab:ablation_act}
\end{table}

%\fontsize{4.6pt}{6.5pt}
\begin{table}[t!]
\centering
\fontsize{4.6pt}{5.5pt} \selectfont
\caption{Ablation for $\alpha$ based on LPIPS distance, $L_{s-t}$ represents LPIPS between source and target. Last row shows LPIPS value between source and target.}
\begin{tabular}{cccc|cc|cc|cc|cc}
\hline
\multicolumn{1}{l}{} & \multicolumn{1}{l}{} & \multicolumn{2}{c|}{\cellcolor[HTML]{60E7F5}\textbf{Sketches ($T_1$)}} & \multicolumn{2}{c|}{\cellcolor[HTML]{60E7F5}\textbf{Female ($T_2$)}} & \multicolumn{2}{c|}{\cellcolor[HTML]{60E7F5}\textbf{Sunglasses ($T_3$)}} & \multicolumn{2}{c|}{\cellcolor[HTML]{60E7F5}\textbf{Male ($T_4$)}} & \multicolumn{2}{c}{\cellcolor[HTML]{60E7F5}\textbf{Babies ($T_5$)}} \\ \cline{3-12} 
$\alpha_{fc}$ & $\alpha_{conv}$ & \cellcolor[HTML]{C0C0C0}FID & \cellcolor[HTML]{C0C0C0}B-LPIPS & \cellcolor[HTML]{C0C0C0}FID & \cellcolor[HTML]{C0C0C0}B-LPIPS & \cellcolor[HTML]{C0C0C0}FID & \cellcolor[HTML]{C0C0C0}B-LPIPS & \cellcolor[HTML]{C0C0C0}FID & \cellcolor[HTML]{C0C0C0}B-LPIPS & \cellcolor[HTML]{C0C0C0}FID & \cellcolor[HTML]{C0C0C0}B-LPIPS \\ \hline
$L_{s-t}$ & \multicolumn{1}{c|}{$L_{s-t}$} & 49.81 & \textbf{\color[HTML]{FE0000} 0.284} & \textbf{\color[HTML]{FE0000} 35.30} & \textbf{\color[HTML]{FE0000} 0.448} & \color[HTML]{000000} 29.91 & \color[HTML]{000000} 0.469 & \textbf{\color[HTML]{FE0000} 34.40} & \textbf{\color[HTML]{FE0000} 0.422} & 76.27 & 0.489 \\
$2 L_{s-t}$ & \multicolumn{1}{c|}{$L_{s-t}/2$} & 47.37 & 0.280 & 38.43 & 0.469 & 27.01 & 0.433 & 39.07 & 0.409 & 50.76 & 0.507 \\
$3 L_{s-t}$ & \multicolumn{1}{c|}{$L_{s-t}/3$} & 46.12 & 0.278 & 38.16 & 0.417 & 26.14 & 0.468 & 42.54 & 0.389 & 48.95 & 0.520 \\
$4 L_{s-t}$ & \multicolumn{1}{c|}{$L_{s-t}/4$} & \textbf{\color[HTML]{FE0000} 45.03} & 0.281 & 38.90 & 0.394 & \textbf{\color[HTML]{FE0000} 25.93} & \textbf{\color[HTML]{FE0000} 0.493} & 42.64 & 0.385 & \textbf{\color[HTML]{FE0000} 43.83} & \cellcolor[HTML]{FFFFFF}{\color[HTML]{FE0000} \textbf{0.526}} \\
$5 L_{s-t}$ & \multicolumn{1}{c|}{$L_{s-t}/5$} & 50.03 & 0.272 & 38.01 & 0.410 & 37.73 & 0.482 & 42.83 & 0.378 & 48.64 & 0.501 \\ \hline
\multicolumn{2}{c|}{$L_{s-t}$ LPIPS distance} & \multicolumn{2}{c|}{0.735} & \multicolumn{2}{c|}{0.253} & \multicolumn{2}{c|}{0.451} & \multicolumn{2}{c|}{0.309} & \multicolumn{2}{c}{0.531} \\ \hline
\end{tabular}
\label{tab:alpha_dist}
\end{table}

%% \fontsize{4.5pt}{6.5pt}
\begin{table}[t!]
\centering
\fontsize{4.5pt}{5.5pt} \selectfont
\caption{Ablation study for the LoRA tensors in mapping and synthesis part.}
\begin{tabular}{c|ccc|cc|cc|cc|cc|cl}
\toprule \hline
\multicolumn{1}{l}{} & \multicolumn{1}{l}{} & \multicolumn{2}{c|}{\cellcolor[HTML]{60E7F5}\textbf{Sketches ($T_1$)}} & \multicolumn{2}{c|}{\cellcolor[HTML]{60E7F5}\textbf{Female ($T_2$)}} & \multicolumn{2}{c|}{\cellcolor[HTML]{60E7F5}\textbf{Sunglasses ($T_3$)}} & \multicolumn{2}{c|}{\cellcolor[HTML]{60E7F5}\textbf{Male ($T_4$)}} & \multicolumn{2}{c|}{\cellcolor[HTML]{60E7F5}\textbf{Babies ($T_5$)}} & \multicolumn{2}{c}{\cellcolor[HTML]{60E7F5}\textbf{Average}} \\ \cline{3-14} 
LoRA FC & LLoRA Conv & \cellcolor[HTML]{C0C0C0}FID & \cellcolor[HTML]{C0C0C0}B-LPIPS & \cellcolor[HTML]{C0C0C0}FID & \cellcolor[HTML]{C0C0C0}B-LPIPS & \cellcolor[HTML]{C0C0C0}FID & \cellcolor[HTML]{C0C0C0}B-LPIPS & \cellcolor[HTML]{C0C0C0}FID & \cellcolor[HTML]{C0C0C0}B-LPIPS & \cellcolor[HTML]{C0C0C0}FID & \cellcolor[HTML]{C0C0C0}B-LPIPS & \multicolumn{1}{l}{\cellcolor[HTML]{C0C0C0}FID} & \cellcolor[HTML]{C0C0C0}B-LPIPS \\ \hline
\checkmark & \multicolumn{1}{c|}{\xmark} & 50.03 & 0.217 & 41.69 & 0.384 & 27.62 & 0.488 & 51.42 & 0.398 & 50.92 & 0.410 & 44.33 & 0.379 \\
\xmark & \multicolumn{1}{c|}{\checkmark} & 55.04 & 0.230 & 38.37 & 0.427 & 46.84 & 0.492 & 52.88 & 0.418 & 76.18 & 0.484 & 53.86 & 0.397 \\
\checkmark & \multicolumn{1}{c|}{\checkmark} & \color[HTML]{FF0000} \textbf{45.56} & \textbf{\color[HTML]{FF0000} 0.280} & \textbf{\color[HTML]{FF0000} 35.30} & \textbf{\color[HTML]{FF0000} 0.448} & \textbf{\color[HTML]{FF0000}25.93} & \cellcolor[HTML]{FFFFFF}{\color[HTML]{FF0000} \textbf{0.493}} & \textbf{\color[HTML]{FF0000} 36.19} & \textbf{\color[HTML]{FF0000} 0.422} & \textbf{\color[HTML]{FF0000}47.20} & \textbf{\color[HTML]{FF0000} 0.527} & \multicolumn{1}{l}{\textbf{\color[HTML]{FF0000}37.91}} & \textbf{\color[HTML]{FF0000} 0.431} \\ \bottomrule
\end{tabular}
\label{tab:ablation_fc_conv}
\end{table}

%\fontsize{3.5pt}{6.5pt}
\begin{table}[t!]
\centering
\fontsize{3.5pt}{5.5pt} \selectfont
\caption{Ablation for $\alpha$ based on LPIPS distance from LSUN-car (used as source dataset for pretraining), and comparison with LFS-GAN.}\label{tab:car}
\begin{tabular}{ll|cc|cc|l|llcccc|l|llcccc}  \toprule
\cline{1-6} \cline{8-13} \cline{15-20}
&  & \multicolumn{2}{c|}{\cellcolor[HTML]{60E7F5}\textbf{Wrecked  cars ($T_1$)}} & \multicolumn{2}{c|}{\cellcolor[HTML]{60E7F5}\textbf{Truck ($T_2$)}} &  &  & \multicolumn{1}{l|}{} & \multicolumn{2}{c|}{\cellcolor[HTML]{60E7F5}\textbf{Wrecked  cars ($T_1$)}} & \multicolumn{2}{c|}{\cellcolor[HTML]{60E7F5}\textbf{Truck ($T_2$)}} &  &  & \multicolumn{1}{l|}{} & \multicolumn{2}{c|}{\cellcolor[HTML]{60E7F5}\textbf{Wrecked  cars ($T_1$)}} & \multicolumn{2}{c}{\cellcolor[HTML]{60E7F5}\textbf{Truck ($T_2$)}} \\ 
$\alpha_{fc}$ & $\alpha_{conv}$ & \cellcolor[HTML]{C0C0C0}FID & \cellcolor[HTML]{C0C0C0}B-LPIPS & \cellcolor[HTML]{C0C0C0}FID & \cellcolor[HTML]{C0C0C0}B-LPIPS &  & $\alpha_{fc}$ & \multicolumn{1}{l|}{$\alpha_{conv}$} & \cellcolor[HTML]{C0C0C0}FID & \multicolumn{1}{c|}{\cellcolor[HTML]{C0C0C0}B-LPIPS} & \cellcolor[HTML]{C0C0C0}FID & \cellcolor[HTML]{C0C0C0}B-LPIPS &  & $\alpha_{fc}$ & \multicolumn{1}{l|}{$\alpha_{conv}$} & \cellcolor[HTML]{C0C0C0}FID & \multicolumn{1}{c|}{\cellcolor[HTML]{C0C0C0}B-LPIPS} & \cellcolor[HTML]{C0C0C0}FID & \cellcolor[HTML]{C0C0C0}B-LPIPS \\ \cline{1-6} \cline{8-13} \cline{15-20} 
$L_{s-t}$ & $L_{s-t}$ & {\color[HTML]{000000} 330.7} & {\color[HTML]{6200C9} {\ul 0.232}} & {\color[HTML]{000000} 221.1} & {\color[HTML]{6200C9} {\ul 0.229}} &  & $L_{s-t}$ & \multicolumn{1}{l|}{$L_{s-t}$} & {\color[HTML]{000000} 330.7} & \multicolumn{1}{c|}{\cellcolor[HTML]{FFFFFF}{\color[HTML]{000000} 0.232}} & {\color[HTML]{000000} 221.1} & \cellcolor[HTML]{FFFFFF}{\color[HTML]{000000} 0.229} &  & $2L_{s-t}$ & \multicolumn{1}{l|}{$L_{s-t}/2$} & {\color[HTML]{000000} 355.4} & \multicolumn{1}{c|}{{\color[HTML]{000000} 0.154}} & {\color[HTML]{000000} 226.2} & {\color[HTML]{000000} 0.201} \\
$2 L_{s-t}$ & $L_{s-t}/2$ & 355.4 & 0.154 & 226.2 & 0.201 &  & $2 L_{s-t}$ & \multicolumn{1}{l|}{$L_{s-t}$} & \cellcolor[HTML]{FFFDFD}{\color[HTML]{6200C9} {\ul 322.5}} & \multicolumn{1}{c|}{0.220} & \cellcolor[HTML]{FFFDFD}{\color[HTML]{000000} 220.2} & 0.204 &  & $2 L_{s-t}$ & \multicolumn{1}{l|}{$L_{s-t}/4$} & 344.8 & \multicolumn{1}{c|}{0.128} & 219.6 & 0.206 \\
$3 L_{s-t}$ & $L_{s-t}/3$ & 356.9 & 0.143 & 230.6 & 0.185 &  & $4 L_{s-t}$ & \multicolumn{1}{l|}{$L_{s-t}$} & 329.5 & \multicolumn{1}{c|}{0.202} & 222.8 & 0.189 &  & $2 L_{s-t}$ & \multicolumn{1}{l|}{$L_{s-t}/8$} & 344.6 & \multicolumn{1}{c|}{0.198} & {\color[HTML]{FF0000} \textbf{214.4}} & 0.215 \\
$4 L_{s-t}$ & $L_{s-t}/4$ & 584.6 & 0.000 & 301.1 & 0.000 &  & $8 L_{s-t}$ & \multicolumn{1}{l|}{$L_{s-t}$} & 337.3 & \multicolumn{1}{c|}{0.180} & 228.1 & 0.180 &  & $2 L_{s-t}$ & \multicolumn{1}{l|}{$L_{s-t}/10$} & \cellcolor[HTML]{FFFFFF}334.8 & \multicolumn{1}{c|}{0.220} & \cellcolor[HTML]{FFFFFF}214.5 & 0.210 \\ \cline{1-6} \cline{8-13} \cline{15-20} 
\multicolumn{2}{c|}{LFS-GAN} & {\color[HTML]{FF0000} \textbf{314.7}} & {\color[HTML]{FF0000} \textbf{0.251}} & {\color[HTML]{6200C9} {\ul 217.5}} & {\color[HTML]{FF0000} \textbf{0.244}} &  & \multicolumn{6}{l|}{} &  & \multicolumn{6}{l}{} \\ \cline{1-6}
\multicolumn{2}{c|}{$L_{s-t}$} & \multicolumn{2}{c|}{0.692} & \multicolumn{2}{c|}{0.813} & \multicolumn{1}{c|}{\cellcolor[HTML]{FFFDFD}} & \multicolumn{6}{l|}{\multirow{-2}{*}{}} &  & \multicolumn{6}{l}{\multirow{-2}{*}{}} \\ \cline{1-6} \cline{8-13} \cline{15-20} \bottomrule

\end{tabular}
\end{table}

\section{Conclusion}
This paper presents a CL and FS image generation approach called CoLoR GAN by integrating LoRA into StyleGAN2. 
To better handle the problem of parameters reduction, we challenged LoRA limits and introduced a LoRA in LoRA (LLoRA) technique for convolutional layers, which further effectively breaks up its weights.
Finally, aware of the LoRA sensitivity to hyper-parameters choice, we provide an empirical study to easily find the best ones.
Our method enables efficient adaptation to target tasks with only few examples while preserving the quality and diversity of the source domain. 
Our experiments demonstrated that our model has competitive performance compared to SOTA but with half parameters and half training iterations.

\section*{Acknowledgements}
This work was funded by Partenariato FAIR (Future Artificial Intelligence Research) - PE00000013, CUP J33C22002830006" funded by the European Union - NextGenerationEU through the Italian MUR within NRRP, project DL-MIG.

\bibliographystyle{splncs04}
\bibliography{bibliography}
\end{document}